\documentclass[times,twocolumn,final,authoryear]{elsarticle}

\usepackage{ycviu}
\usepackage{framed,multirow}

\usepackage{amsmath,amssymb} %
\usepackage{latexsym}
\usepackage[linesnumbered,vlined,ruled]{algorithm2e}

\usepackage{url}
\definecolor{newcolor}{rgb}{.8,.349,.1}
\usepackage{soul}
\usepackage[dvipsnames]{xcolor}

\newcommand{\ie}{\textit{i}.\textit{e}.}
\newcommand{\eg}{\textit{e}.\textit{g}.}
\usepackage{pifont}
\newcommand{\cmark}{\ding{51}}%

\journal{}

\begin{document}

\ifpreprint
  \setcounter{page}{1}
\else
  \setcounter{page}{1}
\fi

\begin{frontmatter}

\title{SSMTL++: Revisiting self-supervised multi-task learning for video anomaly detection}

\author[1]{Antonio \snm{Barbalau}} 
\author[1,2,3]{Radu Tudor \snm{Ionescu}\corref{cor1}}
\cortext[cor1]{Corresponding author: }
\ead{raducu.ionescu@gmail.com}
\author[1,2]{Mariana-Iuliana \snm{Georgescu}}
\author[4,5]{Jacob \snm{Dueholm}}
\author[6]{Bharathkumar \snm{Ramachandra}}
\author[4,5]{Kamal \snm{Nasrollahi}}
\author[3,7]{Fahad Shahbaz \snm{Khan}}
\author[4]{Thomas B. \snm{Moeslund}}
\author[8]{Mubarak \snm{Shah}}

\address[1]{Department of Computer Science, University of Bucharest, 14 Academiei Street, Bucharest 010014, Romania}
\address[2]{SecurifAI, 21D Mircea Voda, Bucharest 030662, Romania}
\address[3]{MBZ University of Artificial Intelligence, Masdar City, Abu Dhabi, UAE}
\address[4]{Department of Architecture, Design, and Media Technology, Aalborg University, Rendsburggade 14, Aalborg 9000, Denmark}
\address[5]{Milestone Systems, Banemarksvej 50C, Br{\o}ndby 2605, Denmark}
\address[6]{Geopipe Inc, 460 W 51st, New York City 10019, NY, US}
\address[7]{Link\"{o}ping University, 581 83 Link\"{o}ping, Sweden}
\address[8]{Center for Research in Computer Vision (CRCV), University of Central Florida, Orlando 32816, FL, US}

\received{1 May 2013}
\finalform{10 May 2013}
\accepted{13 May 2013}
\availableonline{15 May 2013}
\communicated{S. Sarkar}

\begin{abstract}
A self-supervised multi-task learning (SSMTL) framework for video anomaly detection was recently introduced in literature. Due to its highly accurate results, the method attracted the attention of many researchers. In this work, we revisit the self-supervised multi-task learning framework, proposing several updates to the original method. First, we study various detection methods, \eg~based on detecting high-motion regions using optical flow or background subtraction, since we believe the currently used pre-trained YOLOv3 is suboptimal, \eg~objects in motion or objects from unknown classes are never detected. Second, we modernize the 3D convolutional backbone by introducing multi-head self-attention modules, inspired by the recent success of vision transformers. As such, we alternatively introduce both 2D and 3D convolutional vision transformer (CvT) blocks. Third, in our attempt to further improve the model, we study additional self-supervised learning tasks, such as predicting segmentation maps through knowledge distillation, solving jigsaw puzzles, estimating body pose through knowledge distillation, predicting masked regions (inpainting), and adversarial learning with pseudo-anomalies. We conduct experiments to assess the performance impact of the introduced changes. Upon finding more promising configurations of the framework, dubbed SSMTL++v1 and SSMTL++v2, we extend our preliminary experiments to more data sets, demonstrating that our performance gains are consistent across all data sets. In most cases, our results on Avenue, ShanghaiTech and UBnormal raise the state-of-the-art performance bar to a new level.
\end{abstract}


\end{frontmatter}



\section{Introduction}
\label{sec_Intro}

\begin{figure*}[!th]
\begin{center}
  \includegraphics[width=1.0\linewidth]{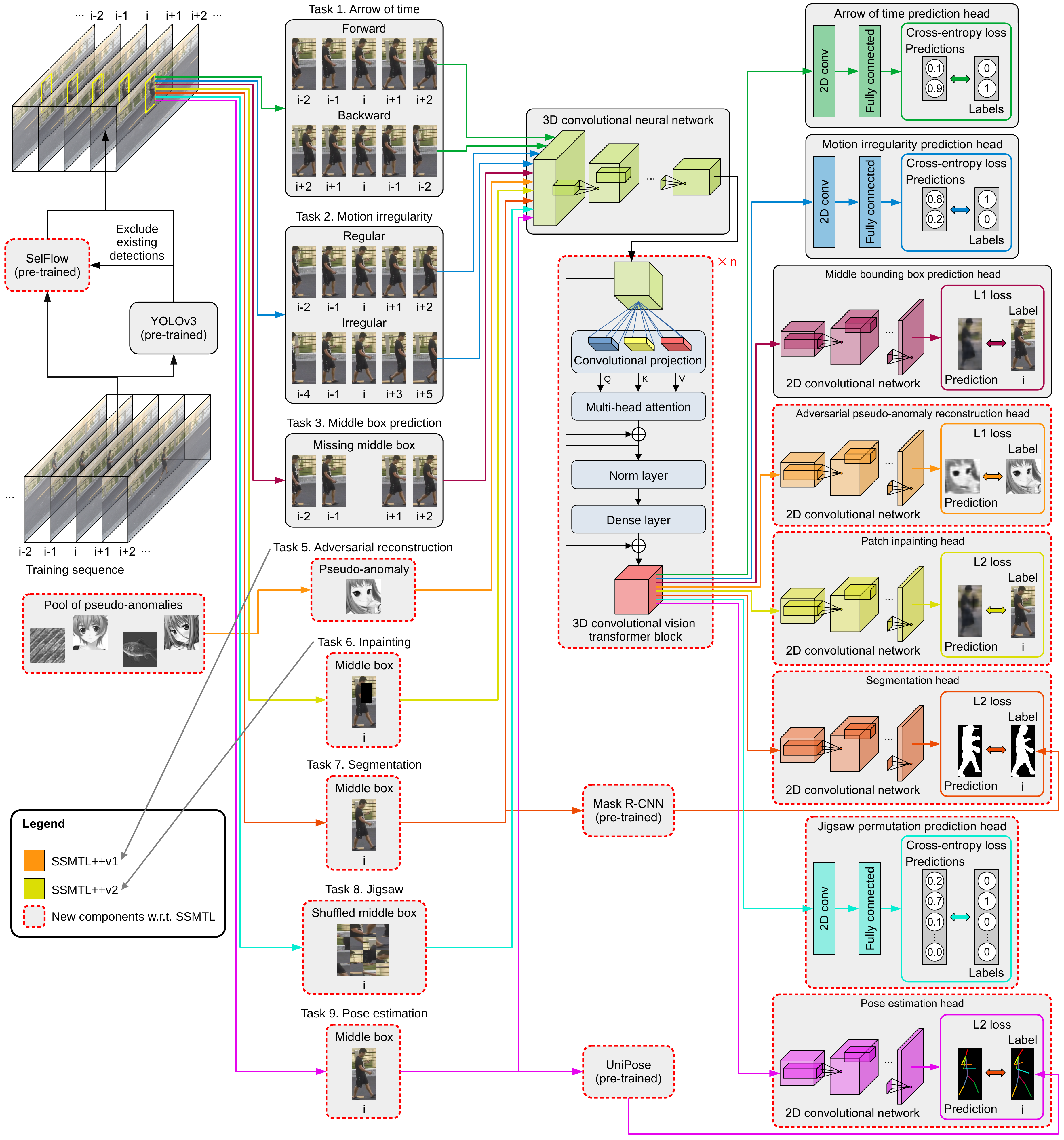}
\end{center}
\vspace{-0.6cm}
  \caption{Overview of the proposed SSMTL++v1 and SSMTL++v2 frameworks. Both SSMTL++v1 and SSMTL++v2 use a hybrid detection method based on YOLOv3 and optical flow, as well as an enhanced backbone formed of a 3D convolutional network followed by a 3D convolutional vision transformer (CvT) \citep{Wu-ICCV-2021}. The first promising combination of tasks (termed SSMTL++v1) is formed of tasks $T_1$ (arrow of time prediction), $T_2$ (motion irregularity prediction), $T_3$ (middle bounding box prediction) and $T_5$ (adversarial reconstruction), while the second promising combination of tasks (termed SSMTL++v2) is formed of tasks $T_1$, $T_2$, $T_3$ and $T_6$ (patch inpainting). Notice that task $T_4$ (knowledge distillation), which was used in the original SSMTL method \citep{Georgescu-CVPR-2021}, is now removed from both SSMTL++v1 and SSMTL++v2. For completeness, the illustrated pipeline also includes tasks $T_7$ (segmentation), $T_8$ (jigsaw) and $T_9$ (pose estimation), which did not lead to performance gains. Our novel components are illustrated inside red dashed contours. Best viewed in color.\vspace{-0.3cm}}
\label{fig_method}
\end{figure*}

Due to its applicability in video surveillance, anomaly detection is an actively studied topic in the video domain, with many recent attempts trying to solve the problem by employing various approaches ranging from outlier detection models \citep{Antic-ICCV-2011,Cheng-CVPR-2015,Cong-CVPR-2011,Dong-Access-2020,Dutta-AAAI-2015,Hasan-CVPR-2016,Ionescu-WACV-2019,Kim-CVPR-2009,Lee-TIP-2019,Li-PAMI-2014,Liu-CVPR-2018,Lu-ICCV-2013,Luo-ICCV-2017,Mahadevan-CVPR-2010,Mehran-CVPR-2009,Park-CVPR-2020,Ramachandra-WACV-2020a,Ramachandra-WACV-2020b,Ramachandra-MVA-2021,Ravanbakhsh-WACV-2018,Ravanbakhsh-ICIP-2017,Ren-BMVC-2015,Sabokrou-IP-2017,Tang-PRL-2020,Wu-TNNLS-2019,Xu-CVIU-2017,Zhao-CVPR-2011,Zhang-PR-2020,Zhang-PR-2016} and weakly-supervised learning frameworks \citep{Feng-CVPR-2021,Purwanto-ICCV-2021,Sultani-CVPR-2018,Tian-ICCV-2021,Zaheer-ECCV-2020,Zhong-CVPR-2019} to supervised open-set methods \citep{Acsintoae-CVPR-2022}. Despite the numerous attempts in solving the problem, video anomaly detection remains a challenging task, especially due to the fact that abnormal events are determined by the context. For example, a pedestrian walking on the sidewalk is a normal event, but a pedestrian who crosses the street far from a crosswalk is an abnormal event (in some countries, pedestrians can even get fines for crossing the street in forbidden areas). Furthermore, since abnormal events are typically rare, it is difficult to collect training data to build fully supervised models. This adds to the difficulty of solving the task. Hence, more research efforts are required towards solving video anomaly detection.

Perhaps one of the most promising directions in video anomaly detection without anomalies at training time is to address the task by learning a self-supervised framework on multiple proxy tasks, which are correlated to anomaly detection, as proposed by \cite{Georgescu-CVPR-2021}. Indeed, \cite{Georgescu-CVPR-2021} introduced a self-supervised multi-task learning (SSMTL) method that learns a set of four proxy tasks using a single 3D convolutional backbone with multiple heads (one head per task), obtaining state-of-the-art performance levels. 

Although the SSMTL model attains very good results, we consider that the framework has a very high potential of obtaining even better results, which can be unearthed by studying and evaluating component variations in depth. 
To this end, we revisit our self-supervised multi-task learning framework \citep{Georgescu-CVPR-2021}, proposing several updates that boost the performance of the method, especially when we combine these updates and generate new frameworks, which we term SSMTL++v1 and SSMTL++v2. 
We identified three important components worth revisiting, namely the object detection method, the multi-task learning backbone architecture, and the proxy tasks. First, we study additional detection methods, \eg~based on detecting high-motion regions using optical flow or background subtraction, since we conjecture the currently used pre-trained YOLOv3 is suboptimal, \eg~objects in motion or objects from unknown classes are never detected. Next, we modernize the 3D convolutional backbone by introducing multi-head self-attention modules, inspired by the recent success of vision transformers \citep{Bertasius-ICML-2021,Dosovitskiy-ICLR-2020,Wu-ICCV-2021}. As such, we alternatively introduce both 2D and 3D convolutional vision transformer (CvT) blocks \citep{Wu-ICCV-2021}. Finally, we study additional proxy tasks, such as predicting segmentation maps through knowledge distillation, solving jigsaw puzzles, estimating body pose through knowledge distillation, predicting masked regions (patch inpainting), and adversarial learning with pseudo-anomalies. Our updates are illustrated inside red dashed contours in Figure~\ref{fig_method}. 

We perform preliminary experiments to determine the impact of introducing the novel components into the SSMTL framework. Through the preliminary experiments, we find two novel and promising combinations (SSMTL++v1 and SSMTL++v2). We evaluate our new frameworks on three benchmark data sets: Avenue \citep{Lu-ICCV-2013}, ShanghaiTech \citep{Luo-ICCV-2017} and UBnormal \citep{Acsintoae-CVPR-2022}. We report considerable performance gains with respect to SSMTL \citep{Georgescu-CVPR-2021}, while also attaining superior results compared to other recent state-of-the-art methods \citep{Astrid-BMVC-2021,Astrid-ICCVW-2021,Bertasius-ICML-2021,Chang-RP-2022,Dong-Access-2020,Doshi-CVPRW-2020a,Doshi-CVPRW-2020b,Georgescu-TPAMI-2021,Gong-ICCV-2019,Ionescu-CVPR-2019,Ionescu-WACV-2019,Ji-IJCNN-2020,Lee-TIP-2019,Li-CVIU-2021,Lin-AAAI-2022,Liu-ICCV-2021,Lu-ECCV-2020,Madan-ICCVW-2021,Nguyen-ICCV-2019,Park-WACV-2022,Park-CVPR-2020,Ramachandra-WACV-2020a,Ramachandra-WACV-2020b,Ristea-CVPR-2022,Sultani-CVPR-2018,Sun-ACMMM-2020,Vu-AAAI-2019,Wang-ACMMM-2020,Wu-TNNLS-2019,Yang-Access-2021,Yu-CVPR-2022,Yu-ACMMM-2020,Yu-TNNLS-2021,Zaheer-CVPR-2022}.

In summary, our main contribution is to augment our preliminary self-supervised multi-task learning framework \citep{Georgescu-CVPR-2021} by improving the detection approach, upgrading the backbone, and introducing new self-supervised proxy tasks. We divide our main contribution into three independent contributions, which are listed in the decreasing order of their importance below:
\begin{itemize}
    \item We study various proxy tasks to be included into the SSMTL framework, finding novel task combinations that produce superior performance levels.
    \item We introduce convolutional vision transformer blocks into the backbone architecture, reporting performance improvements with our stronger backbone.
    \item We introduce additional detection methods into SSMTL to increase the number of detected objects, providing empirical evidence to showcase the benefit of each detection method.
\end{itemize}

\section{Related work}

\noindent
\textbf{Video anomaly detection.}
One dimension of taxonomy divides video anomaly detection methods into those that address single-scene and multi-scene problem formulations. Under this classification, the self-supervised multi-task learning framework treats the multi-scene formulation of the problem, where the training set may contain videos from multiple scenes and anomalies are not expected to be location-dependent. Another dimension categorizes methods into distance-based \citep{Ionescu-CVPR-2019,Ramachandra-WACV-2020b,Ramachandra-MVA-2021,Ravanbakhsh-WACV-2018,Saligrama-CVPR-2012,Smeureanu-ICIAP-2017,Tran-BMVC-2017,Xu-BMVC-2015}, probabilistic \citep{Antic-ICCV-2011,Cheng-CVPR-2015,Feng-NC-2017,Hinami-ICCV-2017,Kim-CVPR-2009,Mehran-CVPR-2009,Wu-CVPR-2010}, reconstruction-based \citep{Astrid-BMVC-2021,Gong-ICCV-2019,Hasan-CVPR-2016,Huang-TNNLS-2022,Huang-TII-2022,Huang-TC-2022,Luo-ICCV-2017,Luo-PAMI-2022,Nguyen-ICCV-2019,Park-CVPR-2020,Ravanbakhsh-ICIP-2017,Ristea-CVPR-2022,Vu-AAAI-2019} and change detection \citep{Giorno-ECCV-2016,Ionescu-ICCV-2017,Liu-BMVC-2018} approaches. As SSMTL is based on multiple self-supervised tasks, it is not possible to place it into only one of these categories. For example, due to the \textit{middle bounding box prediction} task, SSMTL can be viewed as a reconstruction-based approach, where the basic premise is to train a model that reconstructs normal data with higher fidelity as compared to abnormal data, and some composite measure of the reconstruction error subsequently acts as the anomaly score. Similar to several other methods in video anomaly detection \citep{Cheng-CVPR-2015,Feng-NC-2017,Hinami-ICCV-2017,Ionescu-CVPR-2019,Ramachandra-MVA-2021,Vu-AAAI-2019}, SSMTL operates at the object patch level as opposed to at the frame level, but is one of the few reconstruction-based methods to do so. For an extensive treatment of taxonomy in video anomaly detection, we refer the reader to the survey of \cite{Ramachandra-PAMI-2020}. 

Certainly, we consider SSMTL \citep{Georgescu-CVPR-2021} the closest method to the approaches presented in this paper, namely SSMTL++v1 and SSMTL++v2. We underline that the contributions presented in Section~\ref{sec_Intro} also represent the differences with respect to our previous work \citep{Georgescu-CVPR-2021}.

\noindent
\textbf{Multi-task learning.}
As computing devices get faster and more specialized for deep learning applications, benefiting from more memory and processing units, multi-task learning approaches have started gaining popularity, such as with Mask-RCNN \citep{He-ICCV-2017} for object detection and instance segmentation. The basic underlying premise of multi-task approaches is that learning to solve multiple tasks relevant to a primary (target) task is beneficial. When dealing with a problem such as video anomaly detection, where anomalous data is not provided at training time, the primary task cannot be directly supervised; herein lies the motivation for using multi-task learning. Multi-task approaches to video anomaly detection have been used sparsely before \citep{Park-CVPR-2020,Tang-PRL-2020}. However, to the best of our knowledge, SSMTL is the first to propose multi-task learning explicitly and intentionally for generalizing better to the out-of-distribution anomalous patterns in video. Certainly, our current work is based on the same principle.

\noindent
\textbf{Self-supervised learning.}
Self-supervised learning is garnering traction with recent advances showing that, under the right conditions, self-supervised pre-training can outperform fully supervised pre-training in terms of transfer performance in downstream tasks \citep{He-CVPR-2022,He-CVPR-2020}. Self-supervised learning has been widely used before for video anomaly detection. Most reconstruction-based approaches use some form of self-supervised learning. The major approaches include frame-level reconstruction \citep{Hasan-CVPR-2016}, future frame prediction \citep{Dong-Access-2020,Liu-CVPR-2019} or middle frame prediction \citep{Lee-TIP-2019}. The SSMTL framework is however the first to use middle patch prediction at the object level.


\section{Method}
\label{sec_method}

\noindent
\textbf{Original framework.}
In our previous work \citep{Georgescu-CVPR-2021}, we proposed SSMTL, an object-centric framework based on self-supervised and multi-task learning on four proxy tasks. Indeed, the proposed model is trained on three self-supervised tasks and one knowledge distillation task.
The whole architecture is composed of a shared 3D CNN backbone and four independent prediction or reconstruction heads (one per proxy task). The last layer of the shared 3D CNN is global temporal pooling. Therefore, the prediction heads can use 2D convolutions.

The first step of the SSMTL framework is to obtain the object bounding boxes using the YOLOv3 \citep{Redmon-arXiv-2018} object detector. For each object in the frame $i$, a so-called \textit{object-centric temporal sequence} is created by cropping the corresponding bounding box from the frames $\{i - t, ... , i - 1, i, i + 1, ... , i + t\}$.  The \textit{object-centric temporal sequence} is used as input to the 3D CNN.

The first proxy task ($\mathbf{T_1}$) is \textit{predicting the arrow of time}, where the model learns to predict if the temporal sequence is moving forward or backward in time. The second proxy task ($\mathbf{T_2}$) is \textit{predicting motion irregularity}, where the model is trained to predict if the object-centric temporal sequence is cropped from consecutive or intermittent frames, \ie~some frames are skipped in the forward direction to create irregular motion. The third self-supervised proxy task ($\mathbf{T_3}$) is \textit{middle bounding box prediction}, where the middle crop (the bounding box cropped from the frame $i$) is deleted from the temporal sequence and the model is trained to predict the content of the missing bounding box. The fourth proxy task ($\mathbf{T_4}$) is \textit{model distillation}. Here, the model is trained to predict the pre-softmax features of a pre-trained ResNet-50 \citep{He-CVPR-2016} and the class probabilities predicted by the YOLOv3 \citep{Redmon-arXiv-2018} object detector. 

The model is jointly optimized on all four proxy tasks. During inference, the object detector is applied on each frame. Then, for each detected object, the object-centric temporal sequence is created. The temporal sequence is passed through the CNN model, obtaining the output of each prediction head. For $T_1$, the probability that the temporal sequence is moving backward is interpreted as the anomaly score. Similarly, the probability of the temporal sequence to be intermittent is considered as the anomaly score for $T_2$. For $T_3$, the anomaly score is computed as the mean absolute difference between the reconstruction of the middle bounding box predicted by the model and the ground-truth middle bounding box. For $T_4$, only the absolute difference between the class probabilities predicted by the YOLOv3 object detector and those predicted by the model are taken into consideration, saving the time needed to run ResNet-50 during inference. The anomaly score for each object is computed as the average of the anomaly scores given by each prediction head.

\noindent
\textbf{Updates overview.}
We identified three main components that are promising candidates for receiving updates that could positively impact the performance of SSMTL. The first component is the object detection method, which is currently based on a single pre-trained object detector. Increasing the number of detected objects is likely to increase the number of detected anomalies. Hence, we consider adding more detection methods. The second component is the shared backbone architecture. Here, we propose and evaluate two different ways of integrating transformer blocks, which could strengthen the learning capacity of the framework. The last component worth investigating is the set of proxy tasks. \cite{Georgescu-CVPR-2021} showed that the four proxy tasks employed in the original SSMTL framework are useful, but we believe that there are many other proxy tasks that could prove beneficial. We present our updates to these three components in separate sections below.

\subsection{Introducing new detection methods}

The object detector is an important part of the framework because the anomaly analysis is performed only on the detected objects (regions). Thus, if the object detector fails to detect an object of interest (anomalous), the framework will completely miss the respective anomalous event.
In the original SSMTL framework, the YOLOv3~\citep{Redmon-arXiv-2018} object detector was employed. Our first update is to consider the YOLOv5~\citep{Jocher-Z-2022} detector, which is known to outperform YOLOv3. Even though YOLOv5 is supposed to detect the objects more accurately than its previous versions, \eg~YOLOv3, an object detector can only detect a predefined set of classes. However, the set of object classes that can generate an anomaly should not be limited to a fixed number of classes, otherwise we might encounter a frame with objects which the object detector is unable to detect, \eg~a tree that falls on the street, blocking the traffic. Another limitation of object detectors is the inability to detect objects affected by severe motion blur. However, fast moving objects, \eg~a person running, are very likely to cause an anomaly. Due to these limitations, the SSMTL framework can miss such anomalous objects. To alleviate these issues, we propose to detect objects using optical flow or background subtraction, in conjunction with a pre-trained object detector. 

\noindent
{\bf Optical flow.}
On the one hand, we propose to detect new objects belonging to unknown object classes or that are affected by motion blur by applying optical flow. For each frame, we compute the optical flow map with the method proposed by \cite{Liu-CVPR-2019}. We consider that a pixel from the optical flow map is part of a moving object if its magnitude is larger than a threshold. Additionally, we use the YOLO bounding boxes to blackout regions of already detected objects. The resulting connected components are added to the set of detected objects. To eliminate very small objects created by the noise in the optical flow map, we impose a restriction for the width and height of each new object detected through optical flow.

\noindent
{\bf Background subtraction.}
On the other hand, we propose to use background subtraction as a faster alternative to optical flow. We employ a fast and intuitive approach that starts by converting the RGB frames to grayscale. We compute an initial background image for each scene by averaging several consecutive frames. We continuously update the background throughout the video sequence. We subtract the background image from each frame to obtain the foreground objects. Then, we apply a threshold to separate the foreground pixels from the background pixels and clean up the result with a morphological closing operation. As for the optical flow maps, we blackout the regions already detected by YOLO and add the connected components with an area higher than a certain threshold as new objects. 

\subsection{Introducing new backbones}

We extend the original 3D CNN backbone of the SSMTL architecture and introduce a CvT \citep{Wu-ICCV-2021} module formed of multiple sequential transformer blocks. We study two approaches of applying the transformer before and after the global temporal pooling layer, effectively modeling the input as either 2D or 3D. Hence, we name the two approaches 2D CvT and 3D CvT. For the 2D CvT module, we apply average pooling across time to obtain a 2D input of $8\times8$ tokens, before introducing the module. For the 3D CvT module, we first apply the module directly on the output of the shared 3D CNN backbone, the input for the CvT module having $7\times8\times8$ tokens. To make it compatible, we replace the 2D depthwise convolutions from CvT with 3D depthwise convolutions. The output of the 3D CvT module is passed through a global temporal pooling layer, producing the final output. The number of transformer blocks $n$ as well as the number of attention heads $h$ are decided according to the number of proxy tasks, based on a set of preliminary experiments discussed in Section~\ref{sec_prelim}.

We underline that our architecture is designed to cope with inputs of various temporal sizes, which is required to integrate multiple tasks with different temporal input dimensions. This is achieved by following the original architecture proposed by \cite{Georgescu-CVPR-2021}, which has only one temporal pooling layer just before applying the 2D output heads. By using 3D convolutional layers with the padding set to \emph{same}, our backbone is able to process volumes of different temporal sizes, and by using the temporal average pooling just before the 2D heads, we obtain a feature map of the same size, regardless of the temporal input size.

%
%
%
%

\subsection{Introducing new proxy tasks}

Aside from the existing four tasks, we propose a new set of five proxy tasks, as described below. We note that the prediction and decoding heads used for the new tasks are identical to those used for the original four tasks.

\noindent
{\bf $\mathbf{T_5}$: Adversarial reconstruction.}
For the \emph{adversarial reconstruction} of pseudo-anomalies task ($T_5$), we employ an adversarial training procedure based on reversing the gradients, thus performing gradient ascent through our shared backbone, instead of the usual gradient descent. This task was originally introduced by \cite{Georgescu-TPAMI-2021} in the context of video anomaly detection. Similar to \cite{Georgescu-TPAMI-2021}, we attach a decoder to our shared encoder, which learns to reproduce images from an out-of-distribution data set. We consider the same out-of-distribution data set as \cite{Georgescu-TPAMI-2021}, containing flowers \citep{Nilsback-CVPR-2006}, textures \citep{Lazebnik-PAMI-2005}, and ImageNet \citep{Russakovsky-IJCV-2015} classes that do not appear in urban surveillance videos. We optimize the decoder using gradient descent, as usual. However, the shared backbone (encoder) is optimized to model pseudo-anomalies poorly, that is, when updating the network, we reverse the sign of the learning rate for pseudo-anomalies, which is equivalent to performing gradient ascent. This design incentivizes the network to model general patterns poorly, directing the model towards overfitting the provided normal data. Notice that, in the context of anomaly detection, we need a model that does not generalize too well on out-of-distribution data, otherwise it would prevent us from leveraging the reconstruction error as a method for anomaly detection. This is the reason behind applying adversarial training on pseudo-abnormal examples. Different from \cite{Georgescu-TPAMI-2021}, who rely only on this task to subsequently train binary classifiers and predict anomaly scores, we integrate adversarial training as a proxy task into our multi-task learning framework. In addition, we note that our approach is different from the method proposed by \cite{Astrid-BMVC-2021}, since this related method aims to reconstruct unmodified frames from pseudo-abnormal frames without changing the learning procedure, \ie~without reversing the gradients, as we do.

\noindent
{\bf $\mathbf{T_6}$: Patch inpainting.}
The patch inpainting task was introduced by \cite{Pathak-CVPR-2016} in order to learn self-supervised features and improve the accuracy on downstream tasks, such as classification and detection. Our novel contribution consists of applying the patch inpainting task at the object level and leveraging the inpainting error to estimate the abnormality level of the input object. 
Our network is tasked with reconstructing a single image of $64 \times 64$ pixels, out of which, a random patch is cropped out. At test time, for a given input, this procedure is repeated three times in a row, with the output of the previous step serving as input for the current one. At each step, a different random patch is masked out. Finally, we employ the $L_2$ distance between the original image and the final reconstruction to estimate the anomaly level. 

\noindent
{\bf $\mathbf{T_7}$: Segmentation.}
We pose the \emph{segmentation} task as a knowledge distillation task, where the teacher is a pre-trained Mask R-CNN \citep{He-ICCV-2017} and the student is our model. For this task, we attach a decoder to predict the segmentation map. We employ the $L_2$ loss between the predicted map and the ground-truth map to train our student. We use only the middle crop in the temporal sequence as input.

\noindent
{\bf $\mathbf{T_8}$: Jigsaw.}
The \emph{jigsaw} puzzle solving task was originally introduced by \cite{Noroozi-ECCV-2016} to learn CNN features for transfer learning in a self-supervised manner. Different from \cite{Noroozi-ECCV-2016}, we repurpose the task as a proxy task for object-level anomaly detection and integrate it into a multi-task learning setting to detect anomalies in video.
We split each input image into $4\times 4$ patches (puzzle pieces) of $16\times 16$ pixels each and apply a random shuffle of the patches, out of $100$ predefined shuffles. We then task the network to predict the applied shuffle in a multi-way classification setting, using softmax.


\noindent
{\bf $\mathbf{T_9}$: Pose estimation.}
As for the segmentation task, we consider pose estimation as a knowledge distillation task and only use the middle crop in the temporal sequence as input. As teacher, we choose the pre-trained UniPose~\citep{Artacho-CVPR-2020}. The student network is tasked with predicting heatmaps for each body joint, being optimized with the $L_2$ loss.

\subsection{Inference}

During inference, we follow the same simple and straightforward approach as the original framework \citep{Georgescu-CVPR-2021}. We start by detecting the objects in each frame, and then, we extract the object-centric temporal sequence for each detected object. We pass each object-centric sequence $X$ through our neural model, obtaining the score for each proxy task. 
For the arrow of time, motion irregularity and middle bounding box prediction tasks, we apply the same anomaly score interpretation as \cite{Georgescu-CVPR-2021}. More precisely,  we interpret the probability of the temporal sequence to move backward as the anomaly score for the arrow of time proxy task. We consider the probability of the continuous sequence $X$ to be intermittent as the anomaly score for the motion irregularity proxy task. For the middle bounding box prediction task, we use the mean absolute error between the ground-truth object and the reconstructed object as the anomaly score. 

Next, we detail how  we derive the anomaly scores for the newly introduced proxy tasks.
The adversarial reconstruction task ($T_5$) does not contribute to the final anomaly score, being only used to limit the reconstruction power of our model, \ie~to prevent the model from generalizing to out-of-distribution samples.
For the patch inpainting proxy task ($T_6$), we consider the mean absolute difference between the final reconstruction and the target object as the anomaly score. 
For the knowledge distillation tasks, namely the segmentation ($T_7$) and pose estimation ($T_9$) proxy tasks, we interpret the anomaly score as the mean squared error between the teacher's output and our model's output. 
Let $p_{\scriptsize{\mbox{identity}}}$ be the probability predicted by the jigsaw head for the input image to be identically permuted (we do not apply any permutation to the input at test time). The anomaly score for task $T_8$ is given by $1-p_{\scriptsize{\mbox{identity}}}$.

The final anomaly score for each object is the average of the anomaly scores given by the proxy tasks. We underline that we do not take any action to ensure that the value of each loss is within the same range. Following~\cite{Georgescu-CVPR-2021}, for each frame, we next reconstruct a pixel-level anomaly map based on the anomaly score and the location of each detected object. For any two overlapping bounding boxes, we keep the maximum anomaly score for the overlapping region. As a final post-processing step, we apply a 3D mean average filter on the anomaly volume, following~\cite{Giorno-ECCV-2016}. The frame-level anomaly score is given by the maximum score of the pixel-level anomaly map corresponding to each frame. Lastly, we apply a temporal Gaussian filter to smooth the frame-level predictions, following~\cite{Giorno-ECCV-2016}.

\section{Experiments}

\subsection{Data sets}

\noindent
\textbf{Avenue.} The Avenue \citep{Lu-ICCV-2013} data set consists of $16$ training videos containing only normal activity, and $21$ test videos containing both normal and abnormal actions. The resolution of each frame is $360 \times 640$ pixels. The data set is annotated at the frame and pixel levels.

\noindent
\textbf{ShanghaiTech.} The ShanghaiTech Campus \citep{Luo-ICCV-2017} data set is one of the largest data sets for video anomaly detection, containing $437$ videos. The training set contains $330$ videos with normal actions, while the test set consists of $107$ videos with both normal and abnormal events. The resolution of each frame in the data set is $480 \times 856$ pixels. The ShanghaiTech data set is annotated at both frame and pixel levels.

\noindent
\textbf{UBnormal.} The UBnormal \citep{Acsintoae-CVPR-2022} data set is a new supervised open-set benchmark containing abnormal actions in the training videos which are disjoint from the abnormal actions from the test videos. The entire data set has a total of $543$ videos which are divided into $268$ training videos, $64$ validation videos and $211$ test videos. The resolution of the frames can vary, the minimum side of a frame being $720$ pixels. UBnormal is also annotated at both frame and pixel levels. In the experiments, we only use the normal videos to train our framework, preserving its self-supervised nature.

\subsection{Evaluation and implementation details}

\noindent
\textbf{Evaluation metrics.}
We employ the widely-used area under the curve (AUC) computed with respect to the ground-truth frame-level annotations to evaluate detection performance. Following \citep{Acsintoae-CVPR-2022, Georgescu-TPAMI-2021, Ristea-CVPR-2022}, we report both the micro and macro frame-level AUC scores when we compare to other state-of-the-art methods. To evaluate the localization performance of the proposed frameworks, we consider the region-based detection criterion (RBDC) and track-based detection criterion (TBDC) introduced by \cite{Ramachandra-WACV-2020a}. To draw our conclusions after each preliminary experiment, we only look at the more popular micro AUC measure, which we consider the most important due to the sheer number of previous works that have used it \citep{Ramachandra-PAMI-2020}.

\noindent
\textbf{Hyperparameter settings.}
We start from the official implementation of SSMTL\footnote{\url{https://github.com/lilygeorgescu/AED-SSMTL}}. We keep all the original hyperparameters of the framework, established by \cite{Georgescu-CVPR-2021}. For YOLOv3 and YOLOv5, we set the object detection confidence threshold to $0.8$. We use a pre-trained SelFlow~\citep{Liu-CVPR-2019} model to detect objects in motion, forming the objects out of pixels with a motion magnitude higher than $1$. For the SelFlow and background subtraction methods, we eliminate all objects with an area smaller than $1500$ pixels. As \cite{Georgescu-CVPR-2021}, we use temporal sequences of length $7$, resulting in input tensors of $7\times64\times64\times3$ components. Since we remove task $T_4$ in our final framework configurations (SSMTL++v1/v2), we eliminate the hyperparameter $\lambda$ from the original framework, using equal weights for all the remaining proxy tasks.

All models are optimized for $20$ epochs using Adam \citep{Kingma-ICLR-2014} with a learning rate of $\eta=10^{-3}$, keeping the default values for the other hyperparameters of Adam. Depending on the capacity of the backbone architecture, we train the models on mini-batches of $64$, $128$ or $256$ samples. We keep $15\%$ of the training data for validation and choose the model with the lowest validation error on the proxy tasks to be employed on the target task (anomaly detection).

There are some additional hyperparameters for the new proxy tasks added to the model. For the adversarial training with pseudo-anomalies ($T_5$), we adjust the adversarial learning rate to $-0.2 \cdot \eta$ following \citep{McHardy-NAACL-2019}, thus maximizing the loss when an adversarial example is given as input. For the inpainting task ($T_6$), we generate mask patches of random sizes between $4$ and $32$ pixels. The center of each patch is generated using a 2D Gaussian distribution centered in the middle of the input image, having a standard deviation of $20$ in each direction.

\subsection{Preliminary experiments}
\label{sec_prelim}

The preliminary experiments are structured around ablating each of the components proposed in Section \ref{sec_method}, namely the improved detection approach, the transformer-based backbone architecture, and the additional proxy tasks.

\begin{table}[t]
\caption{Results (in terms of recall and frame-level micro AUC) on Avenue while varying the object detection methods of a framework trained on task $T_3$ (middle bounding box prediction).}\label{tab_detections}
\setlength\tabcolsep{5pt}
\begin{center}
\begin{tabular}{|l|c|c|c|}
\hline
{Detection Approach} & \#Objects & Recall & AUC    \\
\hline \hline
YOLOv3               & $111$k & $88.2\%$ & $83.5\%$ \\
YOLOv3 + Optical Flow& $113$k & $91.3\%$ & $89.6\%$ \\
YOLOv3 + Background  & $118$k & $94.8\%$ & $88.4\%$ \\
\hline
YOLOv5               & $105$k & $87.7\%$ & $86.9\%$ \\
YOLOv5 + Optical Flow& $110$k & $90.9\%$ & $89.0\%$ \\
YOLOv5 + Background  & $120$k & $96.1\%$ & $88.2\%$ \\
\hline
\end{tabular}
\end{center}
\end{table}

\begin{figure}[t]
\begin{center}
   \includegraphics[width=1.0\linewidth]{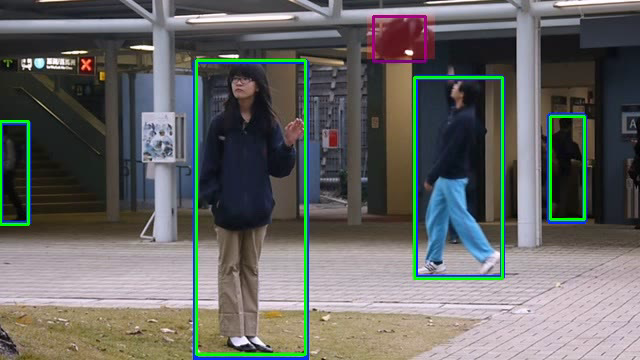}
\end{center}
\vspace{-0.4cm}
   \caption{A test frame from Avenue with bounding boxes found by YOLOv3 (green), YOLOv5 (blue), and optical flow (pink). The ground-truth anomalous region is covered by a translucent red color. Optical flow is able to pick up anomalous objects missed by pre-trained object detectors (possibly due to motion blur). Best viewed in color.}
\label{fig:detections_02}
\end{figure}

\subsubsection{Experimenting with new detection methods}

We first experiment with various detection methods, presenting the results on the Avenue data set in Table~\ref{tab_detections}. To evaluate each detector, we report the total number of detections in the test set together with the recall calculated with respect to the ground-truth annotations provided by \cite{Ramachandra-WACV-2020a}.  On the one hand, the recall measure gives an insight about the usefulness of the object detections, before having to train and test the entire anomaly detection framework. Thus, the recall of the object detector only represents an upper bound for the recall of the anomaly detection framework. On the other hand, the frame-level AUC reported in Table \ref{tab_detections} is based on the true positive and false positive rates of the anomaly detection framework, and is not necessarily proportional to the recall. Hence, higher recall does not automatically point to higher AUC.

We observe that optical flow adds a few additional detections for a better recall, while background subtraction adds a higher number of detections, leading to the highest recall score. However, the AUC score is more relevant for the anomaly detection task.
The original SSMTL framework uses the YOLOv3 detector, which leads to a micro AUC of $83.5\%$. YOLOv5 alone seems to output better object detections, providing a performance gain of $3.4\%$ over YOLOv3. When we add optical flow detections, the performance increases by significant margins for both YOLOv3 and YOLOv5. A typical scenario is shown in Figure \ref{fig:detections_02}, where a backpack is missed by both YOLOv3 and YOLOv5 object detectors, likely due to motion blur. The backpack is however detected via optical flow. 

Interestingly, there seems to be a much higher gain by combining YOLOv3 and optical flow detections. When we introduce the detections obtained via background subtraction, we again observe considerable performance gains, but not as high as compared to optical flow. This is caused by some of the added detections being labeled as anomalies by mistake, generating a higher false positive rate. This shows that the recall measure can be used as a rough guidance towards evaluating the impact of the object detection pipeline, whereas the true anomaly detection capability of the network is accurately quantified by the AUC measure, as a compromise between the true positive and false positive rates.
Hence, we conclude that optical flow is a better choice than background subtraction. Since the combination of YOLOv3 and optical flow gives the best micro AUC ($89.6\%$), we continue with this detection approach in the subsequent experiments.

\begin{table}[t]
\caption{Results (in terms of the frame-level micro AUC) on Avenue while changing the backbone architecture for a model trained on one task ($T_3$ -- middle bounding box prediction).}\label{tab_backbone_1task}
\begin{center}
\begin{tabular}{|l|l|c|}
\hline
{Tasks} & {Backbone} & {AUC} \\
\hline\hline
$T_3$               & 3D CNN            & $89.6\%$ \\
$T_3$               & 3D CNN + 2D CvT   & $90.1\%$ \\
$T_3$               & 3D CNN + 3D CvT   & $90.6\%$ \\
\hline
\end{tabular}
\end{center}
\end{table}

\subsubsection{Experimenting with new backbones}

\noindent
{\bf Old versus new backbone trained on one proxy task.}
In our second experiment, we study the effect of changing the backbone architecture from a pure 3D CNN to one based on transformer blocks. For this empirical study, we select only one task (middle bounding box reconstruction) and perform the experiments on the Avenue data set, just as before. We report the corresponding results in Table~\ref{tab_backbone_1task}. Since the model learns only one task, we choose the shallow and narrow \citep{Georgescu-CVPR-2021} configuration for the 3D CNN. By introducing 2D CvT blocks after performing average pooling across time, we observe a gain of $0.5\%$ in terms of the micro AUC. We notice a higher gain ($1\%$) upon introducing the 3D transformer module prior to the pooling across time. We keep the backbone based on the 3D CvT for the following experiments.

\begin{table}[t]
\caption{Results (in terms of the frame-level micro AUC) on Avenue while varying the number of transformer blocks (from 1 to 3) and the number of attention heads (from 6 to 18). All models are trained on three tasks: $T_1$ (arrow of time), $T_2$ (motion irregularity), $T_3$ (middle bounding box prediction).}\label{tab_architecture_size}
\begin{center}
\begin{tabular}{|c|c|c|c|}
\hline
\multirow{2}{*}{{Blocks}} & \multicolumn{3}{|c|}{{Heads}} \\
\cline{2-4}
                                & {6} & {12} & {18} \\
\hline
\hline
{1} & $91.6\%$ & $90.1\%$ & $91.4\%$ \\
{2} & $92.4\%$ & $92.0\%$ & $90.5\%$ \\
{3} & $91.1\%$ & $92.5\%$ & $92.2\%$ \\
\hline
\end{tabular}
\end{center}
\end{table}

\begin{table}[t]
\caption{Results (in terms of the frame-level micro AUC) on Avenue while changing the backbone architecture for a model trained on three tasks ($T_1$ -- arrow of time, $T_2$ -- motion irregularity, $T_3$ -- middle bounding box prediction).}\label{tab_backbone_3task}
\begin{center}
\begin{tabular}{|l|l|c|}
\hline
{Tasks} & {Backbone} & {AUC} \\
\hline\hline
$T_1$+$T_2$+$T_3$   & 3D CNN            & $90.7\%$ \\
$T_1$+$T_2$+$T_3$   & 3D CNN + 3D CvT   & $92.5\%$ \\
\hline
\end{tabular}
\end{center}
\end{table}

\noindent
{\bf Adjusting the new backbone to more proxy tasks.}
Next, we aim to determine if the performance gains brought by the new transformer-based backbone are consistent when introducing more tasks. We underline that for the original backbone, \cite{Georgescu-CVPR-2021} increased the depth and width of the architecture along with the number of tasks. In a similar manner, we study how the number of transformer blocks (from 1 to 3) and the number of attention heads (from 6 to 18) influences the performance of the framework when we switch from one task (middle bounding box reconstruction) to the following three tasks: $T_1$ (arrow of time prediction), $T_2$ (motion irregularity prediction), $T_3$ (middle bounding box reconstruction). We present the results with various depths and widths on Avenue in Table~\ref{tab_architecture_size}. The empirical results indicate that the best configuration is to use 3 blocks with 12 attention heads each.

\noindent
{\bf Old versus new backbone trained on three tasks.}
Upon finding the optimal architecture in the context of multi-task learning, we now compare the deep+wide 3D CNN with the new backbone based on 3D CNN + 3D CvT with 3 blocks and 12 attention heads. Both models are trained on the first three proxy tasks for a fair comparison. We report the corresponding results in Table~\ref{tab_backbone_3task}. We observe that introducing more tasks increases the gap between the old and new backbones, by up to $1.8\%$. We keep the configuration based on 3 blocks and 12 attention heads for the remaining experiments.

\begin{table}[t]
\caption{Results (in terms of the frame-level micro AUC) on Avenue while varying the proxy tasks for a model based on the 3D CNN + 3D CvT architecture.}\label{tab:alone_proxy_tasks}
\begin{center}
\begin{tabular}{|l|c|}
\hline
{Tasks}                            & {AUC} \\
\hline\hline
$T_1$                       & $83.6\%$ \\
$T_2$                       & $83.4\%$ \\
$T_3$                       & $83.5\%$ \\
$T_4$                       & $73.7\%$ \\
$T_5$                       & - \\
$T_6$                       & $84.3\%$ \\
$T_7$                       & $69.1\%$ \\
$T_8$                       & $50.5\%$ \\
$T_9$                       & $81.5\%$ \\
\hline
\end{tabular}
\end{center}
\end{table}

\begin{table}[t]
\caption{Results (in terms of the frame-level micro AUC) on Avenue while changing the proxy tasks for a model based on the 3D CNN + 3D CvT architecture.}\label{tab:proxy_tasks}
\begin{center}
\begin{tabular}{|l|c|c|}
\hline
{Tasks}                                 & Backbone  & {AUC} \\
\hline\hline
$T_1$+$T_2$+$T_3$                       & shared    & $92.5\%$ \\
$T_1$+$T_2$+$T_3$+$T_4$                 & shared    & $92.0\%$ \\
$T_1$+$T_2$+$T_3$+$T_5$                 & shared    & $93.7\%$ \\
$T_1$+$T_2$+$T_3$+$T_6$                 & shared    & $91.6\%$ \\
$T_1$+$T_2$+$T_3$+$T_7$                 & shared    & $90.7\%$ \\
$T_1$+$T_2$+$T_3$+$T_8$                 & shared    & $90.2\%$ \\
$T_1$+$T_2$+$T_3$+$T_9$                 & shared    & $91.1\%$ \\
$T_1$+$T_2$+$T_3$+$T_4$+$T_5$           & shared    & $89.8\%$ \\
$T_1$+$T_2$+$T_3$+$T_4$+$T_6$           & shared    & $89.5\%$ \\
$T_1$+$T_2$+$T_3$+$T_5$+$T_6$           & shared    & $90.1\%$ \\
$T_1$+$T_2$+$T_3$+$T_4$+$T_5$+$T_6$     & shared    & $89.6\%$ \\
$T_1$+$T_2$+$T_3$+$T_4$+$T_5$+$T_6$     & separate   & $90.5\%$ \\
\hline
\end{tabular}
\end{center}
\end{table}

\subsubsection{Experimenting with new tasks}

\noindent
{\bf Individual proxy task learning.}
In Table~\ref{tab:alone_proxy_tasks}, we present results with individual proxy tasks on the Avenue data set. With the exception of task $T_8$ (jigsaw), the proposed tasks obtain comparable results with the original proxy tasks used by SSMTL. In addition, we emphasize that $T_5$ (adversarial reconstruction) is not applicable as a standalone proxy task (it is only meant to be used in conjunction with other tasks). Hence, task $T_5$ does not contribute to the anomaly score, and there are no results to report for task $T_5$. In summary, the only task that fails to work sufficiently well in detecting anomalies is $T_8$. We believe this happens because the model is likely focusing on the background patterns, which are repetitive across object bounding boxes, to solve the jigsaw puzzles.

\noindent
{\bf Multi-task learning.}
In Table~\ref{tab:proxy_tasks}, we present results with various task combinations on Avenue. First, we underline that the knowledge distillation task ($T_4$) from the original framework is not immediately compatible with the addition of detections based on optical flow or background subtraction, since these detections do not have an assigned object class, unlike the YOLOv3 detections. Moreover, assigning classes to these detections is not trivial, as they sometimes include a single object part or multiple objects. To this end, the experiments conducted so far do not include task $T_4$. However, we can introduce a new class of objects that comprises all the optical flow detections. As shown in Table~\ref{tab:proxy_tasks}, this solution is suboptimal, leading to a slight performance drop from $92.5\%$ to $92.0\%$. 

\begin{figure}[t]
\begin{center}
   \includegraphics[width=0.8\linewidth]{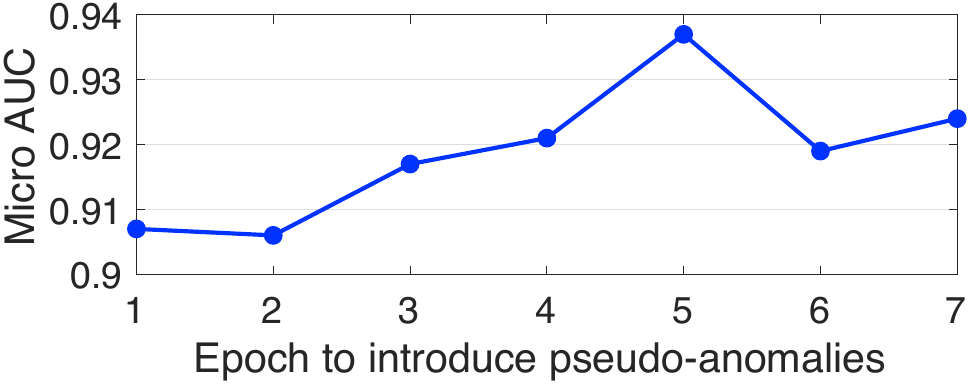}
\end{center}
\vspace{-0.4cm}
   \caption{Frame-level AUC scores showing the effect of introducing pseudo-anomalies at different epochs while training SSMTL++v1 on Avenue.}
\label{fig:epochs}
\end{figure}

\begin{table*}[!th]
\caption{Comparison of the proposed frameworks (SSMTL++v1 and SSMTL++v2) with the original SSMTL \citep{Georgescu-CVPR-2021} as well as other state-of-the-art methods on the Avenue data set. The top three scores for each metric are highlighted in \textbf{\color{red}red bold} (top method), {\color{ForestGreen}green} (second best) and {\color{blue} blue} (third best).}\label{tab_avenue}
\begin{center}
\begin{tabular}{|c|l|c|c|c|c|}
\hline
\multirow{2}{*}{{Year}} & \multirow{2}{*}{Method} & \multicolumn{2}{c|}{AUC} & \multirow{2}{*}{{RBDC}} & \multirow{2}{*}{{TBDC}}  \\
\cline{3-4}
&& {Micro} & {Macro} && \\
\hline
\hline
\multirow{7}{*}{{2019}}&\cite{Gong-ICCV-2019} & $83.3\%$  & - & - & - \\
&\cite{Ionescu-CVPR-2019} & $87.4\%$ & $90.4\%$ & $15.8\%$ & $27.0\%$ \\
&\cite{Ionescu-WACV-2019} & $88.9\%$ & - & - & -  \\
&\cite{Lee-TIP-2019} & $90.0\%$ & - & -  & - \\
&\cite{Nguyen-ICCV-2019} & $86.9\%$  & - & - & - \\
&\cite{Vu-AAAI-2019} & $71.5\%$ & - & - & - \\
&\cite{Wu-TNNLS-2019} & $86.6\%$  & - & - & - \\
\hline
\multirow{10}{*}{{2020}}
&\cite{Dong-Access-2020} & $84.9\%$  & - & -  & - \\
&\cite{Doshi-CVPRW-2020a,Doshi-CVPRW-2020b} & $86.4\%$ & - & - & - \\
&\cite{Ji-IJCNN-2020} & $78.3\%$ & - & - & - \\
&\cite{Lu-ECCV-2020} & $85.8\%$ & - & - & - \\
&\cite{Park-CVPR-2020} & $88.5\%$ & - & -  & - \\
&\cite{Ramachandra-WACV-2020a} & $72.0\%$  & - & $35.8\%$  & $80.9\%$ \\
&\cite{Ramachandra-WACV-2020b} & $87.2\%$  & - & $41.2\%$ & $78.6\%$ \\
&\cite{Sun-ACMMM-2020} & $89.6\%$  & - & - & - \\
&\cite{Wang-ACMMM-2020} & $87.0\%$  & - & - & - \\
&\cite{Yu-ACMMM-2020} & $89.6\%$ & - & -  & - \\ 
\hline 
\multirow{10}{*}{{2021}}
&\cite{Astrid-BMVC-2021} & $84.9\%$ & - & - & - \\
&\cite{Astrid-ICCVW-2021} & $87.1\%$ & - & - & - \\
&\cite{Chang-RP-2022} &  $87.1\%$ & - & - & - \\
&\cite{Georgescu-TPAMI-2021} & $92.3\%$ & $90.4\%$ & ${\color{ForestGreen}65.1\%}$ & $66.9\%$ \\  
&\cite{Madan-ICCVW-2021} & $88.6\%$ & - & - &  - \\  
&\cite{Li-CVIU-2021} & $88.8\%$ & - & - & - \\ 
&\cite{Liu-ICCV-2021} & $91.1\%$ & $\mathbf{\color{red}{93.5\%}}$ & $41.1\%$ & ${\color{ForestGreen}86.2\%}$ \\
&\cite{Yang-Access-2021} & $88.6\%$ & - & - & - \\
&\cite{Yu-TNNLS-2021} & $90.2\%$ & -  & -  & - \\
\cline{2-6}
& SSMTL \citep{Georgescu-CVPR-2021} & $91.5\%$ & ${\color{blue}91.9\%}$  & $57.0\%$  & $58.3\%$ \\
\hline 
\multirow{8}{*}{{2022}}
& \cite{Georgescu-TPAMI-2021}+\cite{Ristea-CVPR-2022} & ${\color{ForestGreen}92.9\%}$ & ${\color{blue}91.9\%}$ & $\mathbf{\color{red}66.0\%}$ & $64.9\%$ \\
& \cite{Lin-AAAI-2022} & $90.3\%$ & - & - & - \\
& \cite{Liu-CVPR-2018} + \cite{Ristea-CVPR-2022} & $87.3\%$ & $84.5\%$ & $20.1\%$ & $62.3\%$ \\
& \cite{Liu-ICCV-2021} + \cite{Ristea-CVPR-2022} & $90.9\%$ & $92.2\%$ & ${\color{blue}62.3\%}$ & $\mathbf{\color{red}89.3\%}$ \\
& \cite{Park-WACV-2022} & $85.3\%$ & - & - & - \\
& \cite{Yu-CVPR-2022} & ${\color{blue}92.8\%}$ & - & - & - \\
\cline{2-6}
& SSMTL++v1 (ours) & $\mathbf{\color{red}93.7\%}$ & $91.7\%$ & $40.9\%$ & $82.1\%$ \\
& SSMTL++v2 (ours) & $91.6\%$ & ${\color{ForestGreen}92.5\%}$ & $47.8\%$ & ${\color{blue}85.2\%}$ \\
\hline
\end{tabular}
\end{center}
\end{table*}

\begin{table*}[!th]
\caption{Comparison of the proposed frameworks (SSMTL++v1 and SSMTL++v2) with the original SSMTL \citep{Georgescu-CVPR-2021} as well as other state-of-the-art methods on the ShanghaiTech data set. The top three scores for each metric are highlighted in \textbf{\color{red}red bold} (top method), {\color{ForestGreen}green} (second best) and {\color{blue}blue} (third best).} \label{tab_shanghai}
\begin{center}
\begin{tabular}{|c|l|c|c|c|c|}
\hline
\multirow{2}{*}{{Year}} & \multirow{2}{*}{Method} & \multicolumn{2}{c|}{AUC} & \multirow{2}{*}{{RBDC}} & \multirow{2}{*}{{TBDC}}  \\
\cline{3-4}
&& {Micro} & {Macro} && \\
\hline
\hline
\multirow{3}{*}{{2019}}&\cite{Gong-ICCV-2019} & $71.2\%$ & - & - & - \\
&\cite{Ionescu-CVPR-2019} & $78.7\%$ & $84.9\%$ & $20.7\%$ & $44.5\%$ \\
&\cite{Lee-TIP-2019} & $76.2\%$  & - & - & -\\
\hline
\multirow{7}{*}{{2020}}
&\cite{Dong-Access-2020} & $73.7\%$  & - & - & - \\
&\cite{Doshi-CVPRW-2020a,Doshi-CVPRW-2020b} & $71.6\%$ & - & - & - \\
&\cite{Lu-ECCV-2020} & $77.9\%$  & - & - & - \\
&\cite{Park-CVPR-2020} & $70.5\%$  & - & - & - \\
&\cite{Sun-ACMMM-2020} & $74.7\%$ & - & - & - \\
&\cite{Wang-ACMMM-2020} & $79.3\%$ & - & - & - \\
&\cite{Yu-ACMMM-2020} & $74.8\%$  & - & - & - \\ 
\hline 
\multirow{9}{*}{{2021}}
&\cite{Astrid-BMVC-2021} & $76.0\%$ & - & - & - \\
&\cite{Astrid-ICCVW-2021} & $73.7\%$ & - & - & - \\
&\cite{Chang-RP-2022} & $73.7\%$ & - & - & - \\
&\cite{Georgescu-TPAMI-2021} & $82.7\%$ & $89.3\%$ & $41.3\%$ & $78.8\%$ \\  
&\cite{Madan-ICCVW-2021} & $74.6\%$ &  - & - & - \\  
&\cite{Li-CVIU-2021} & $73.9\%$ & - & - & - \\
&\cite{Liu-ICCV-2021} & $76.2\%$ & - & - & - \\
&\cite{Yang-Access-2021} & $74.5\%$ & - & - & - \\
\cline{2-6}
& SSMTL \citep{Georgescu-CVPR-2021} & $82.4\%$ & $89.3\%$  & $42.8\%$  & $83.9\%$ \\
\hline 
\multirow{8}{*}{{2022}}
& \cite{Georgescu-TPAMI-2021} + \cite{Ristea-CVPR-2022} & ${\color{ForestGreen}83.6\%}$ & ${\color{blue}89.5\%}$ & $40.6\%$ & $83.5\%$ \\
& \cite{Liu-CVPR-2018} + \cite{Ristea-CVPR-2022} & $74.5\%$ & $82.9\%$ & $18.5\%$ & $60.2\%$ \\
& \cite{Liu-ICCV-2021} + \cite{Ristea-CVPR-2022} & $75.5\%$ & $83.7\%$ & ${\color{ForestGreen}45.5\%}$ & ${\color{ForestGreen}84.5\%}$ \\
& \cite{Park-WACV-2022} & $72.2\%$ & - & - & - \\
& \cite{Yu-CVPR-2022} & $72.1\%$ & - & - & - \\
& \cite{Zaheer-CVPR-2022} & $79.6\%$ & - & - & - \\
\cline{2-6}
& SSMTL++v1 (ours) & ${\color{blue}82.9\%}$ & ${\color{ForestGreen}89.8\%}$ & ${\color{blue}43.2\%}$ & ${\color{blue}84.1\%}$ \\
& SSMTL++v2 (ours) & $\mathbf{\color{red}83.8\%}$ & $\mathbf{\color{red}90.5\%}$ & $\mathbf{\color{red}47.1\%}$ & $\mathbf{\color{red}85.6\%}$ \\
\hline
\end{tabular}
\end{center}
\end{table*}

Next, we experiment with independently adding our new proxy tasks to the first three tasks, to assess the impact of each new proxy task on the performance of the whole framework. Among the new proxy tasks, we find tasks $T_5$ (adversarial reconstruction of pseudo-anomalies) and $T_6$ (patch inpainting) as the most promising. Note that tasks $T_7$ and $T_9$ do not fail at generalizing for anomaly detection when used by themselves, but only in conjunction with the other tasks.

\noindent
{\bf Introducing more tasks.}
We next attempt to combine five or six proxy tasks together, considering the most promising options. However, the results indicate significant performance drops when jointly optimizing the framework on five or more tasks. Although adding $T_5$ and $T_6$ separately improves performance (as mentioned above), their combination does not seem to achieve the same gains. Our first assumption for explaining the lower results is that the backbone needs a higher capacity to cope with the larger number of tasks. We tried to increase its capacity, without obtaining any performance gains. We also tried to use a separate backbone for each task, which seems to be somewhat useful, but not enough to surpass our best performing combination of tasks ($T_1$, $T_2$, $T_3$ and $T_5$). Another explanation for the poor results with five tasks or more is that some combinations of tasks depending on different loss functions and loss magnitudes are simply harder to optimize jointly.

\noindent
{\bf When to introduce adversarial reconstruction.}
We underline that task $T_5$ is not introduced right from the beginning, as the network needs some time to converge on the other tasks before adversarial training is enabled. As confirmed by Figure~\ref{fig:epochs}, it is worth waiting for a few epochs before enabling task $T_5$, the optimal starting point being epoch $5$.

\subsubsection{Chosen SSMTL++ configurations}

For the final comparison with the existing state-of-the-art methods, we choose two of our most promising models. Our first combination of tasks (SSMTL++v1) is formed of tasks $T_1$, $T_2$, $T_3$ and $T_5$. Our second combination of tasks (SSMTL++v2) is formed of tasks $T_1$, $T_2$, $T_3$ and $T_6$. We recall that both SSMTL++v1 and SSMTL++v2 use a hybrid object detection method based on YOLOv3 and optical flow, as well as an enhanced backbone (3D CNN + 3D CvT).

\subsection{Comparison with the state of the art}

\noindent
{\bf Results on Avenue.}
We present the comparative results of SSMTL++v1 and SSMTL++v2 versus the state-of-the-art methods on the Avenue data set in Table~\ref{tab_avenue}. Compared to SSMTL, we observe that SSMTL++v1 and SSMTL++v2 attain better micro AUC, macro AUC and TBDC scores, but the new models register drops in terms of RBDC. The high gains ($+23.8\%$ and $+26.9\%$) in terms of TBDC are caused by introducing the new detections obtained by optical flow, which generate longer and consistent object tracks. Unfortunately, some of the new object detections are labeled as anomalous by mistake, increasing the false positive rate and causing important drops in terms of RBDC. However, the TBDC gains outweigh the RBDC drops. Moreover, SSMTL++v1 attains the highest micro AUC ($93.7\%$) among all models.

\noindent
{\bf Results on ShanghaiTech.}
We present the results of the comparative study conducted on ShanghaiTech in Table~\ref{tab_shanghai}. We observe that both SSMTL++v1 and SSMTL++v2 obtain consistent improvements over SSMTL \citep{Georgescu-CVPR-2021} across all metrics. Remarkably, SSMTL++v2 attains the top performance on each metric, surpassing all other approaches. At the same time, SSMTL++v1 shares the second and third places (depending on the metric) with two state-of-the-art frameworks (\cite{Georgescu-TPAMI-2021} and \cite{Liu-CVPR-2018}) which were recently enhanced with self-supervised predictive convolutional attentive blocks (SSPCAB) \citep{Ristea-CVPR-2022}.

\begin{table*}[t]
\caption{Comparison of the proposed frameworks (SSMTL++v1 and SSMTL++v2) with the original SSMTL \citep{Georgescu-TPAMI-2021} as well as other state-of-the-art methods on the UBnormal data set. The top three scores for each metric are highlighted in \textbf{\color{red}red bold} (top method), {\color{ForestGreen}green} (second best) and {\color{blue} blue} (third best).}\label{tab_ubnormal}
\begin{center}
\begin{tabular}{|l|c|c|c|c|}
\hline
\multirow{2}{*}{Method} & \multicolumn{2}{c|}{AUC} & \multirow{2}{*}{{RBDC}} & \multirow{2}{*}{{TBDC}}  \\
\cline{2-3}
& {Micro} & {Macro} && \\
\hline
\hline
\cite{Sultani-CVPR-2018} (pre-trained)  & $49.5\%$ & $77.4\%$ & $<\!0.01\%$ & $<\!0.01\%$ \\
\cite{Sultani-CVPR-2018} (fine-tuned)  & $50.3\%$ & $76.8\%$ & $<\!0.01\%$ & $<\!0.01\%$ \\
\cite{Bertasius-ICML-2021} (1/32 rate)  & $\mathbf{\color{red}68.5\%}$ & $80.3\%$ & $0.04\%$ & $0.05\%$ \\
\cite{Bertasius-ICML-2021} (1/8 rate)  & ${\color{ForestGreen}64.1\%}$ & $75.4\%$ & $0.04\%$ & $0.05\%$ \\
\cite{Bertasius-ICML-2021} (1/4 rate)  & $61.9\%$ & $75.4\%$ & $0.04\%$ & $0.06\%$ \\
\cite{Georgescu-TPAMI-2021}  & $59.3\%$ & $84.9\%$ & ${\color{blue}{21.91\%}}$ & $53.44\%$ \\
\cite{Georgescu-TPAMI-2021}+UBnormal  & $61.3\%$ & ${\color{blue}85.6\%}$ & ${\color{ForestGreen}25.43\%}$ & ${\color{blue}56.27\%}$ \\
\hline
SSMTL \cite{Georgescu-CVPR-2021} & $55.4\%$ & $84.5\%$  & $19.71\%$  & $55.80\%$ \\
\hline
SSMTL++v1 (ours) & ${\color{blue}62.1\%}$ & $\mathbf{\color{red}86.5\%}$ & $\mathbf{\color{red}25.63\%}$ & $\mathbf{\color{red}63.53\%}$ \\
SSMTL++v2 (ours) & $56.0\%$ & ${\color{ForestGreen}85.9\%}$ & $20.33$\%& ${\color{ForestGreen}57.76\%}$ \\
\hline
\end{tabular}
\end{center}
\end{table*}

\noindent
{\bf Results on UBnormal.}
As the UBnormal \citep{Acsintoae-CVPR-2022} benchmark is very new, the number of existing results is relatively small. Nevertheless, we showcase the results of the comparative study conducted on UBnormal in Table~\ref{tab_ubnormal}. As for the ShanghaiTech data set, we notice that both SSMTL++v1 and SSMTL++v2 surpass the original SSMTL method. Furthermore, SSMTL++v1 establishes new state-of-the-art levels for three metrics (macro AUC, RBDC and TBDC), while SSMTL++v2 is the second best method for two of the metrics (macro AUC and TBDC). Although the TimeSformer of \cite{Bertasius-ICML-2021} seems very good at determining if a video frame is abnormal or not, the model is not able to localize anomalies inside frames. The TBDC and RBDC differences in favor of our models outweigh the lower micro AUC scores compared to TimeSformer.

\begin{figure}[t]
\begin{center}
   \includegraphics[width=1.0\linewidth]{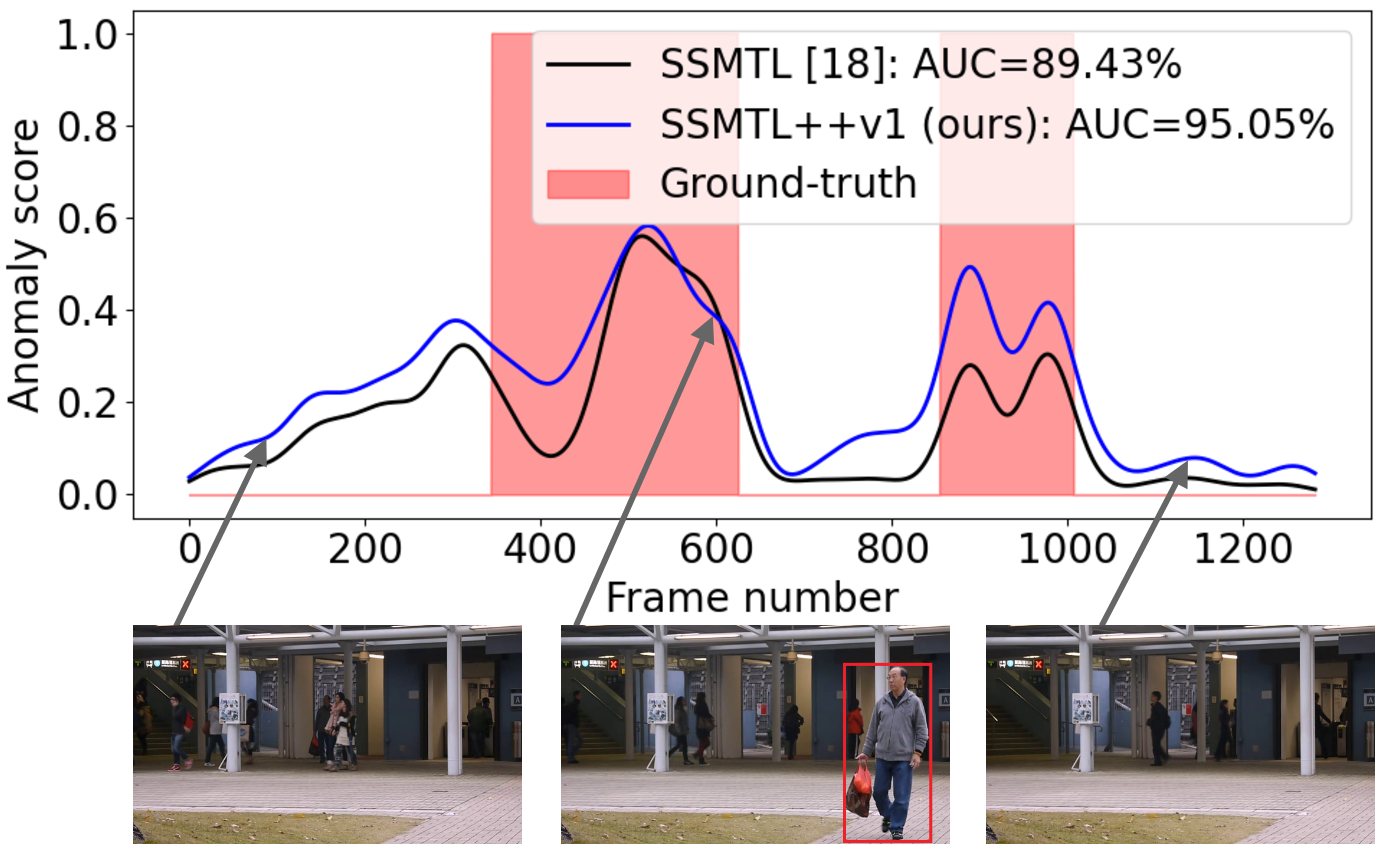}
\end{center}
\vspace{-0.4cm}
   \caption{Comparing the anomaly scores (on the vertical axis) of SSMTL \citep{Georgescu-CVPR-2021} and SSMTL++v1 on the frames (on the horizontal axis) of test video 06 from Avenue. Best viewed in color.}
\label{fig:avenue}
\end{figure}

\begin{figure}[t]
\begin{center}
   \includegraphics[width=1.0\linewidth]{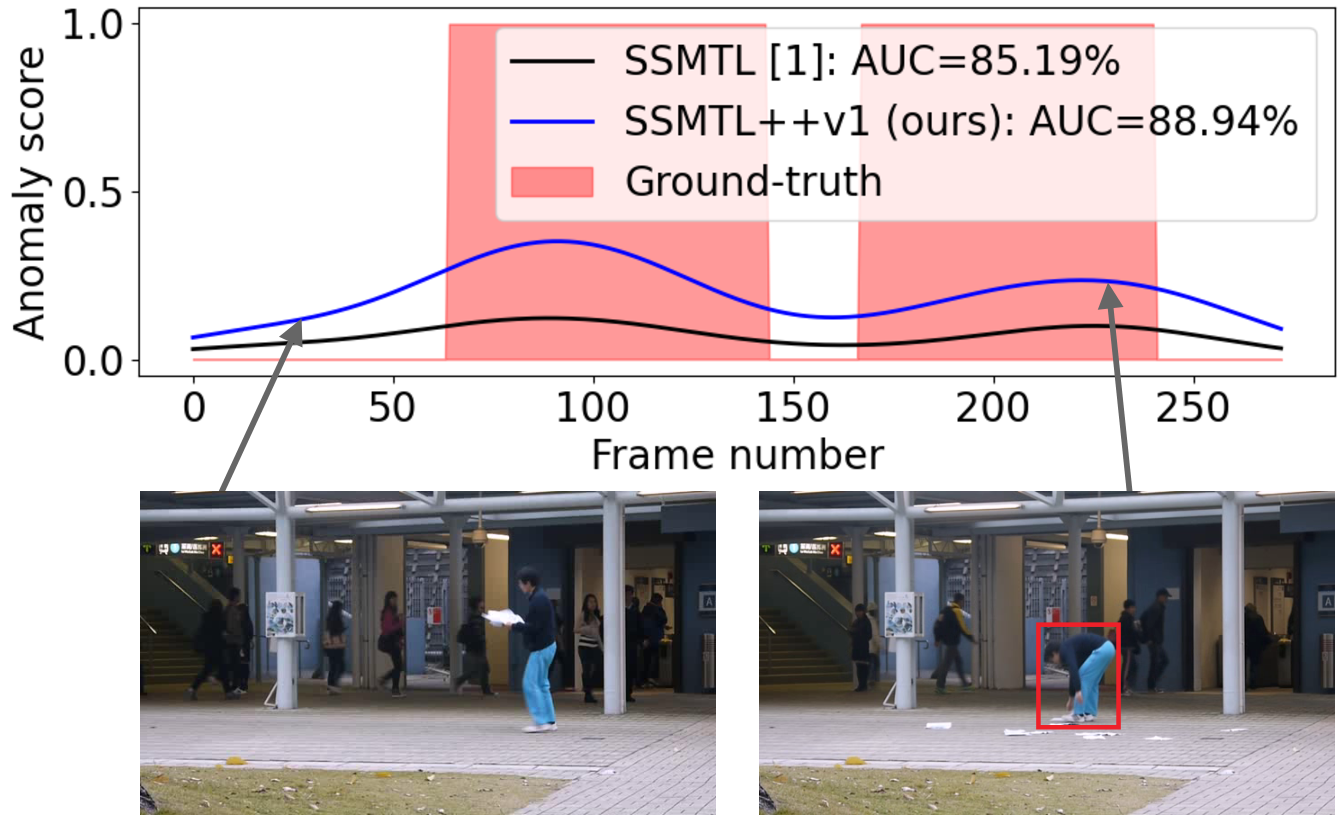}
\end{center}
\vspace{-0.4cm}
   \caption{Comparing the anomaly scores (on the vertical axis) of SSMTL \citep{Georgescu-CVPR-2021} and SSMTL++v1 on the frames (on the horizontal axis) of test video 20 from Avenue. Best viewed in color.}
\label{fig:avenue2}
\end{figure}

\begin{figure}[t]
\begin{center}
   \includegraphics[width=1.0\linewidth]{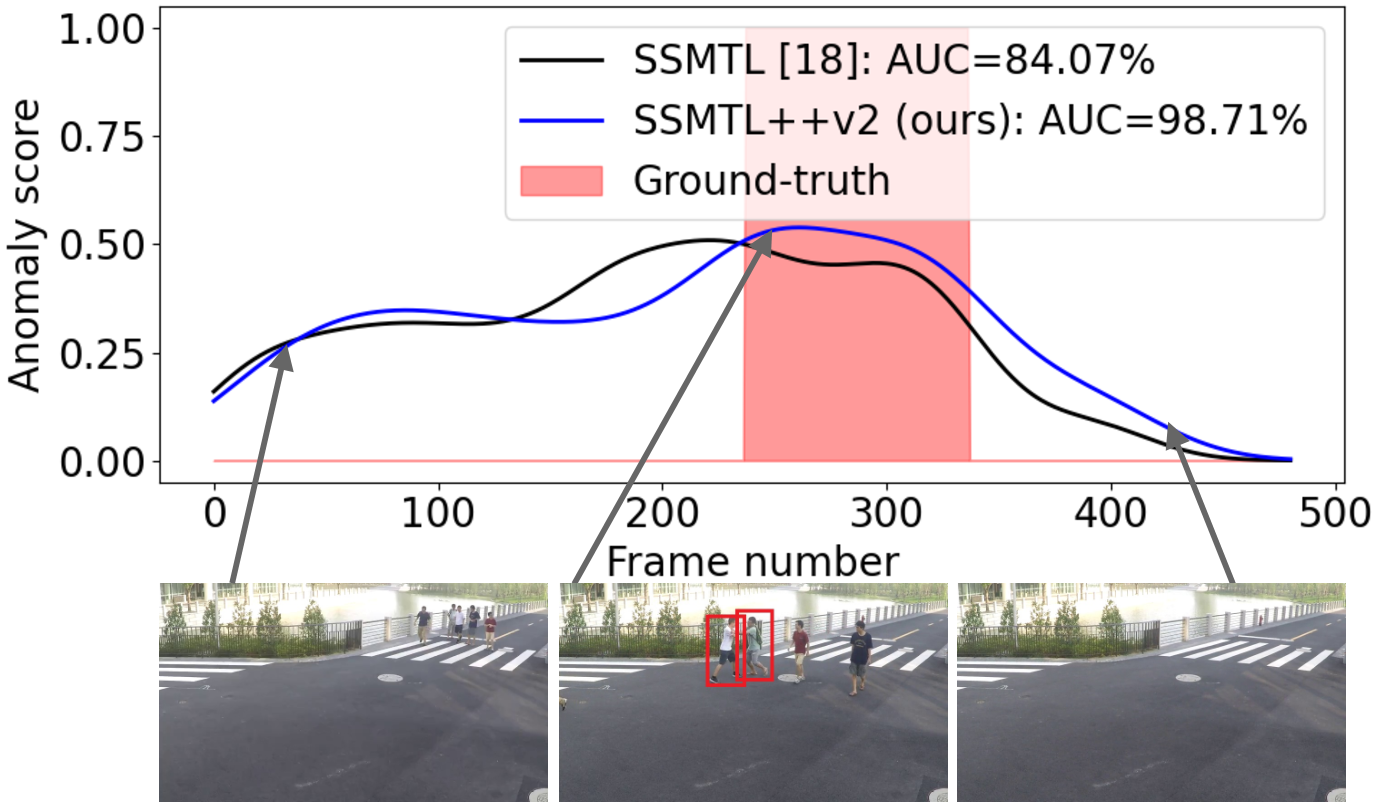}
\end{center}
\vspace{-0.4cm}
   \caption{Comparing the anomaly scores (on the vertical axis) of SSMTL \citep{Georgescu-CVPR-2021} and SSMTL++v2 on the frames (on the horizontal axis) of test video 07\_0049 from ShanghaiTech. Best viewed in color.}
\label{fig:shanghai}
\end{figure}

\begin{figure}[t]
\begin{center}
   \includegraphics[width=1.0\linewidth]{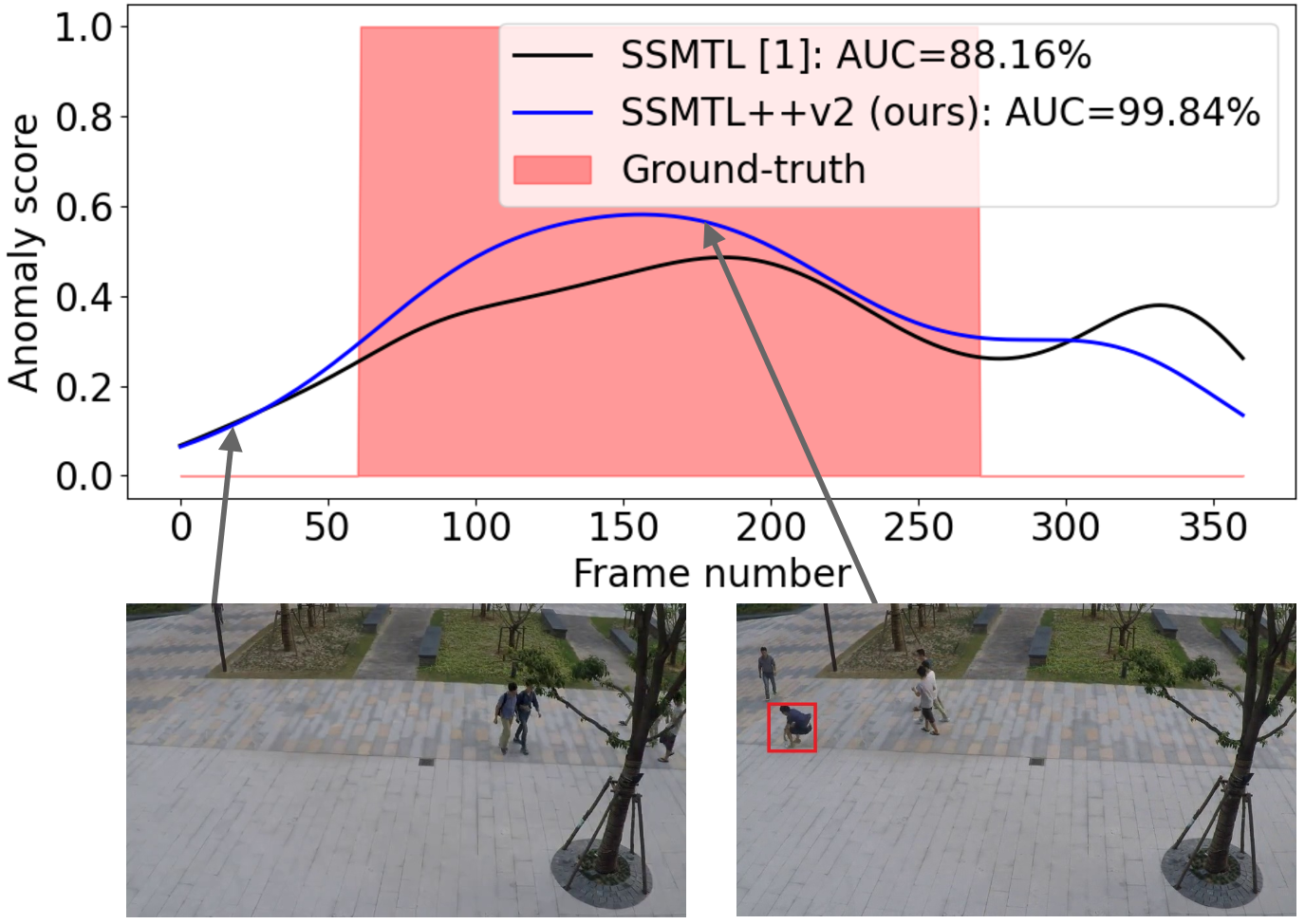}
\end{center}
\vspace{-0.4cm}
   \caption{Comparing the anomaly scores (on the vertical axis) of SSMTL~\citep{Georgescu-CVPR-2021} and SSMTL++v2 on the frames (on the horizontal axis) of test video 04\_0013 from ShanghaiTech. Best viewed in color.}
\label{fig:shanghai2}
\end{figure}

\noindent
{\bf Qualitative analysis.}
We present a few test cases to assess the quality of the predicted anomaly scores for SSMTL and SSMTL++v1/v2 with respect to the ground-truth. 
Upon analyzing Figure \ref{fig:avenue}, we can observe that the AUC gains brought by SSMTL++v1 over SSMTL are higher than $5\%$ on test video 06 from Avenue. The person labeled as anomalous by SSMTL++v1 is walking in the wrong direction. Similarly, in Figure \ref{fig:avenue2}, we can easily notice that the AUC gains brought by SSMTL++v1 over SSMTL are higher than $4\%$ on test video 20 from Avenue. The person labeled as anomalous by SSMTL++v1 is throwing and gathering papers on the ground.
Looking at Figure \ref{fig:shanghai}, we notice that SSMTL++v2 is outperforming SSMTL by a significant margin ($+14\%$) on test video 07\_0049 from ShanghaiTech. SSMTL++v2 labels two humans as anomalous because they are fighting. Figure \ref{fig:shanghai2} also shows that SSMTL++v2 is outperforming SSMTL by a significant margin ($+11\%$) on test video 04\_0013 from ShanghaiTech. SSMTL seems to activate on the last 50 frames (with indexes higher than 310) in the video, producing a false positive event, while SSMTL++v2 does not trigger an anomaly for the same event. SSMTL++v2 outputs higher anomaly scores only for the person jumping, which is labeled as a true anomaly.

\noindent
{\bf Final remarks on the experiments.}
With a few exceptions, SSMTL++v1 and SSMTL++v2 attain very high performance levels, generally surpassing the competing methods across different data sets and metrics. In the majority of cases, our new frameworks reach new state-of-the-art levels. We thus conclude that our updates brought to the SSMTL framework are noteworthy.

\subsection{Ablation study}

\begin{table} 
\caption{Results (in terms of the frame-level micro AUC) on Avenue, while ablating various components of the SSMTL++v1 and SSMTL++v2 framework versions.}\label{tab:ablation_study}
\begin{center}
\begin{tabular}{|c|c|c|c|c|c|}
\hline
\multicolumn{2}{|c|}{Detection Approach}    & \multirow{2}{*}{CvT}       & \multicolumn{2}{|c|}{Extra Task}  & \multirow{2}{*}{AUC}\\
\cline{1-2}
\cline{4-5}
YOLOv3      & Optical Flow                  &           & $\;T_5\;$         & $T_6$             & \\
\hline\hline
\cmark      &                               &           &               &                   & $89.3\%$ \\   
\cmark      & \cmark                        &           &               &                   & $90.7\%$ \\  
\cmark      & \cmark                        & \cmark    &               &                   & $92.5\%$ \\ 
\hline
\cmark      & \cmark                        &           & \cmark        &                   & $90.2\%$ \\
\cmark      &                               & \cmark    & \cmark        &                   & $88.8\%$ \\
\cmark      & \cmark                        &  \cmark   & \cmark        &                   & $93.7\%$ \\
\hline
\cmark      & \cmark                        &           &               & \cmark            & $90.4\%$ \\
\cmark      &                               & \cmark    &               & \cmark            & $88.7\%$ \\
\cmark      & \cmark                        &  \cmark   &               & \cmark            & $91.6\%$ \\
\hline
\end{tabular}
\end{center}
\end{table}

In Table \ref{tab:ablation_study}, we present ablation results on the Avenue data set in terms of frame-level micro AUC, for our SSMTL++v1 and SSMTL++v2 framework versions. We start from the baseline configuration based on YOLOv3 as detection method, 3D CNN as backbone architecture, and $T_1$, $T_2$ and $T_3$ as proxy tasks. In the first part of Table \ref{tab:ablation_study}, we assess the impact of the novel framework components that are used by both SSMTL++v1 and SSMTL++v2, namely the new object detection approach and the new backbone architecture. The ablation results specific to SSMTL++v1 are presented next, followed by the ablation results specific to SSMTL++v2.

\noindent
{\bf Ablation of common components.} We observe that adding the optical flow detections to the YOLOv3 detentions improves the baseline performance by $1.4\%$, from $89.3\%$ to $90.7\%$. Moreover, integrating the CvT backbone architecture further increases the performance to $92.5\%$. Based on the aforementioned results, we conclude that both framework modifications (the upgraded detection approach and the new backbone architecture) improve the framework.

\noindent
{\bf Ablation specific to SSMTL++v1.}
We investigate the influence of task $T_5$ on the performance of the framework in different settings, by removing the optical flow detection approach, or the CvT backbone. When we remove the CvT backbone, the performance drops from $93.7\%$ to $90.2\%$, once again showing the importance of using CvT. We observe a similar effect when removing optical flow.  
The highest performance is obtained when all the components are in place (optical flow, CvT and task $T_5$), each of them bringing improvements to the SSMTL++v1 framework.
 
\noindent
{\bf Ablation specific to SSMTL++v2.}
As for SSMTL++v1, we investigate the effect of proxy task $T_6$ in different scenarios, ablating each component of the SSMTL++v2 framework. By alternatively removing the optical flow and CvT components, the performance decreases by more than $1.2\%$ in terms of frame-level micro AUC, revealing that each framework component is important in obtaining higher performance.

\noindent
{\bf Summary of ablation experiments.}
The ablation results indicate that each component of the SSMTL++v1 and SSMTL++v2 frameworks plays a key role in obtaining higher performance and going beyond the current state-of-the-art results.

\subsection{Running time}

\noindent 
{\bf Training time.} We train the proposed methods on a single GeForce GTX 3090 GPU with $24$ GB of VRAM, using the same setup as for SSMTL. In Table \ref{tab_training_time}, we report the running times required for the preprocessing steps (performed only once, before training) as well as the actual training, on each data set. On our machine, the training time per epoch is roughly $16$ minutes for Avenue, $3$ hours for ShanghaiTech, and $1$ hour and $15$ minutes for UBnormal. Noting that all models are trained for $20$ epochs, it results that the largest training time is $60$ hours (on ShanghaiTech). Moreover, since the preprocessing steps based on YOLOv3 and SelFlow are executed only once per data set, the time required for preprocessing is significantly lower than the time needed for training.
In addition, we underline that training time differences between SSMTL++v1/v2 and SSMTL are negligible. 

\begin{table}[t]
\caption{Preprocessing and training times (in hours and minutes) for SSMTL++v1/v2. The reported running times were measured on a GeForce GTX 3090 GPU with 24 GB of VRAM.}\label{tab_training_time}
\setlength{\tabcolsep}{3.7pt}
\begin{center}
\begin{tabular}{|l|l|r|r|r|}
\hline
{Stage} & {Component}   & {Avenue} & {Shanghai} & {UBnormal} \\
\hline
\hline
\multirow{2}{*}{Preprocessing} & YOLOv3              & 04m  & 1h 04m    & 28m\\   
&SelFlow             & 09m & 2h 32m    &  1h 05m \\
\hline
Training & SSMTL++       & 5h 20m  & 60h 00m & 25h 00m \\ 
\hline
All & All & 5h 33m & 63h 36m & 26h 33m \\ 
\hline
\end{tabular}
\end{center}
\end{table}

\noindent 
{\bf Inference time.}
In Table~\ref{tab_time}, we compare the inference times of the original SSMTL \citep{Georgescu-CVPR-2021} and other state-of-the-art models \citep{Georgescu-TPAMI-2021,Gong-ICCV-2019,Park-WACV-2022,Liu-CVPR-2018,Park-CVPR-2020} with the inference times of SSMTL++v1 and SSMTL++v2.

Regarding the running time, our main changes of the original framework (SSMTL) \citep{Georgescu-CVPR-2021} introduce additional processing steps: $(i)$ in the object detection phase due to optical flow, $(ii)$ in the forward pass through the backbone due to the appended transformer blocks, and $(iii)$ in the prediction phase of SSMTL++v2 due to the additional inpainting task ($T_6$), as it requires 3 passes through the model. We note that the pseudo-anomalies task ($T_5$) of SSMTL++v1 is only used during training, thus not having any influence on the inference time.

YOLOv3 \citep{Redmon-arXiv-2018}, which is used by both SSMTL and SSMTL++v1/v2 methods, takes nearly $0.84$ seconds to process a mini-batch of $64$ frames, thus running at about $72$ frames per second (FPS). The optical flow is obtained using the pre-trained SelFlow \citep{Liu-CVPR-2019}, running at $30$ FPS on mini-batches of $32$ frames. 

The SSMTL++v1 network processes one object-centric temporal sequence in $12$ milliseconds (ms), without batching. For an object-centric approach, we can naturally batch the objects detected in each frame, resulting in a processing time of $3$ ms per object-centric temporal sequence for a mini-batch of $7$ samples, corresponding to the average number of detections per frame in the Avenue test set. SSMTL++v2 processes an object-centric temporal sequence in $4$ ms with an identical mini-batch size.

SSMTL can process the video frames from the Avenue test set at about $50$ FPS. Due to the introduction of optical flow in the object detection phase and the deeper backbone, the inference times for SSMTL++v1 and SSMTL++v2 decrease to about $15$ FPS, considering a sequential processing pipeline on a single thread. However, running each SSMTL++ version on two threads in parallel increases the speed to about $20$ FPS, while still using a single GPU. Hence, both SSMTL++v1 and SSMTL++v2 are fast enough to process the video in real-time. The reported running times were measured on a GeForce GTX 3090 GPU with $24$ GB of VRAM.

In Table~\ref{tab_time}, we also include the running times of other recent methods \citep{Georgescu-TPAMI-2021,Gong-ICCV-2019,Liu-CVPR-2018,Park-CVPR-2020,Park-WACV-2022}, for which the running times were measured on the same hardware configuration. We consider that the running times of our models are still competitive, both SSMTL++v1 and SSMTL++v2 being capable of real-time processing. Furthermore, we note that the faster methods obtain considerably lower performance levels, \eg~the methods proposed in~\citep{Liu-CVPR-2018,Park-CVPR-2020,Park-WACV-2022} yield micro AUC scores that are more than $10\%$ lower compared with SSMTL++v1 and SSMTL++v2 on ShanghaiTech, our largest benchmark.

\begin{table}[t]
\caption{Running time (in terms of FPS) for SSMTL++v1/v2 versus SSMTL and other state-of-the-art models \citep{Liu-CVPR-2018,Gong-ICCV-2019,Park-CVPR-2020,Georgescu-TPAMI-2021,Park-WACV-2022}. The reported running times were measured on a GeForce GTX 3090 GPU with 24 GB of VRAM.}\label{tab_time}
\begin{center}
\begin{tabular}{|l|c|}
\hline
{Method}                         & {FPS} \\

\hline\hline
 
FFP \citep{Liu-CVPR-2018} & 133 \\   
MemAE  \citep{Gong-ICCV-2019}  & 42 \\
MNAD  \citep{Park-CVPR-2020} & 56 \\
Background Agnostic \citep{Georgescu-TPAMI-2021} & 18 \\
FastAno \citep{Park-WACV-2022} & 195 \\ 

\hline
SSMTL \citep{Georgescu-CVPR-2021} & 50.0 \\  
SSMTL++v1                        & 20.2 \\  
SSMTL++v2                        & 18.8 \\  
\hline
\end{tabular}
\end{center}
\end{table}

Our main bottleneck is the optical flow framework. We note that the focus of this work has been on precision in terms of anomaly detection capabilities, without much effort being dedicated towards optimizing the running time. Alternatively, faster object detectors can be applied, as well as using the faster background subtraction ($55$ FPS), at a small cost of precision, if inference time is a higher priority.

\section{Conclusion}

In this work, we revisited the self-supervised multi-task learning framework introduced by \cite{Georgescu-CVPR-2021}, proposing a series of updates that boost the performance of the original method to new state-of-the-art levels. We provided empirical evidence for several beneficial updates. First, we showed that using optical flow along with YOLOv3 to obtain object detections is very useful in finding more objects. Second, we obtained additional performance gains by integrating 3D convolutional multi-head attention blocks into the backbone architecture. Furthermore, we showed that the \emph{adversarial training on pseudo-anomalies} and \emph{patch inpainting} tasks are well-correlated to anomaly detection, leading to performance improvements of the multi-task learning pipeline. Interestingly, these new proxy tasks are useful when replacing the original knowledge distillation task ($T_4$), rather than being jointly added as additional proxy tasks. 

Noting that models trained on more than 5 tasks seem to underperform, in future work, we aim to study more ways to learn from as many tasks as possible, which, at least in principle, should lead to even better results.

\section*{Acknowledgments}

The authors thank reviewers for their valuable feedback, which led to significant improvements of the manuscript.

This work was supported by a grant of the Romanian Ministry of Education and Research, CNCS - UEFISCDI, project number PN-III-P2-2.1-PED-2021-0195, within PNCDI III. This work has also been funded by the Milestone Research Programme at AAU, and by SecurifAI.

\bibliographystyle{model2-names}
\bibliography{refs}


\end{document}